# EnergyPlus Room Simulator

*Manuel* Weber[1*], *Philipp* Bogdain[1], *Sophia* Viktoria Weißenberger[1], *Diana* Marjanovic[1], *Katharina* Sammet[1], *Jan* Vellmer[1], *Farzan* Banihashemi[2], and *Peter* Mandl[1]

[1]HM Munich University of Applied Sciences, Munich, Germany
[2]Technical University of Munich, Munich, Germany

**Abstract.** Research towards energy optimization in buildings heavily relies on building-related data such as measured indoor climate factors. While data collection is a labor- and cost-intensive task, simulations are a cheap alternative to generate datasets of arbitrary sizes, particularly useful for data-intensive deep learning methods. In this paper, we present the tool *EnergyPlus Room Simulator*, which enables the simulation of indoor climate in a specific room of a building using the simulation software EnergyPlus. It allows to alter room models and simulate various factors such as temperature, humidity, and CO2 concentration. In contrast to manually working with EnergyPlus, this tool enhances the simulation process by offering a convenient interface, including a user-friendly graphical user interface (GUI) as well as a REST API. The tool is intended to support scientific, building-related tasks such as occupancy detection on a room level by facilitating fast access to simulation data that may, for instance, be used for pre-training machine learning models.

## 1 Introduction

The building sector has a large impact on climate change, as buildings are responsible for nearly one third of the world's energy consumption [1]. To address this environmental challenge, energy optimization in the context of buildings has emerged as a critical research topic. However, energy optimization depends on various factors, including building construction as well as occupant behavior, and research relies on data, either collected from real buildings or obtained from simulation. EnergyPlus is a well-known open-source simulation software widely adopted in research and practice. To increase access to EnergyPlus for this purpose, we developed the tool *EnergyPlus Room Simulator*. It is designed to (1) facilitate the simulation process for users who are not proficient in simulating with EnergyPlus, and (2) to enable automated room simulations with an arbitrary programming language. For this, the tool consists of a frontend and backend component, offering (1) a graphical user interface (GUI) facilitating the step-by-step creation of simulations as well as (2) a REST API supporting all operational steps required to simulate a room or a number of different rooms. Besides the simulation, the tool allows manipulating room models in the EnergyPlus-specific idf format and provides a 3D visualization of the final room (see Figure 1). The tool is publicly available: source code and documentation are published in a GitHub repository[†].

---

\* Corresponding author: manuel.weber@hm.edu
[†] https://github.com/CCWI/EP-Room-Simulator/

The remainder of this paper is organized as follows: First, we address the motivation for this work, in terms of application cases in research, then we introduce some related simulation tools. This is followed by an outline of the architecture of the room simulator and its functionality.

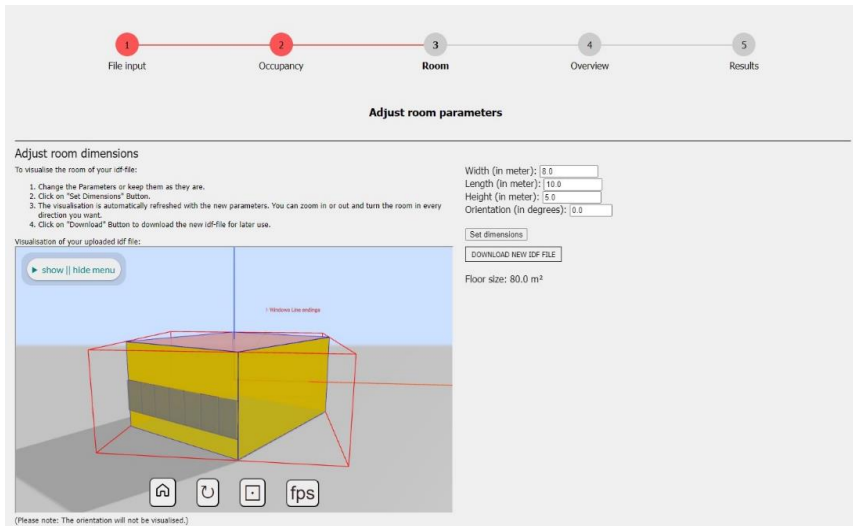

**Fig. 1.** EnergyPlus Room Simulator (Screenshot).

## 2 Motivation

The *EnergyPlus Room Simulator* supports scientific, building-related tasks such as optimizing energy performance or indoor environmental quality, or detecting room-level occupancy or occupant behavior. One exemplary application is the deep learning approach to occupancy detection from $CO_2$ values discussed in [2]. The work addresses the combination of real-world and simulation data to reduce the required amount of collected ground truth data. The evaluation indicates that pre-training with simulation data may reduce the amount of data to be annotated by half and, in addition to this, contributes to a more robust model. In a subsequent study, it was demonstrated that simulated $CO_2$ data performs comparably to real-world data in the context of deep transfer learning for occupancy detection [3]. Similarly, simulation data may also enhance the prediction of other occupant-related information, e.g., the prediction of window-opening behavior. For this task, Banihashemi et al. [4] apply long short-term memory (LSTM) models and dense neural networks on environmental data. They argue that also in this context, transfer learning allows for reducing required ground truth data, which is hard to collect on a large scale, and hence should be investigated in future work. One data source for transfer learning is simulation data. The proposed tool allows to conveniently generate such data for further application in research.

## 3 Related Work

Several works have already presented tools to generate building-related data. Chen et al. [5], for instance, have built an agent-based occupancy simulator to generate stochastic occupancy data. This data may be used to replace static schedules in building performance simulations. Another type of simulator deals with co-simulation, aiming to test building control strategies or reinforcement learning algorithms alongside with a building energy simulation.

This is addressed by *COBS* [6], a python platform for co-simulations with EnergyPlus. Regarding an upfront simulation to generate potential training data for neural networks, including indoor environmental data, Chaudhary et al. [7] have developed the python tool *synconn_build*. It is integrated with EnergyPlus as well, and allows running simulations with randomized inputs to multiple control features. In contrast to *synconn_build*, we focus on altering room geometry, instead of limiting manipulation to operating conditions. In addition to this, we also intend to offer a simple GUI that does not necessarily rely on user knowledge towards EnergyPlus or programming.

## 4 Architecture and Components

Figure 2 illustrates the architecture of the software system. The frontend and backend are based on two independent python flask servers. The frontend sends HTTP requests to a REST API provided by the backend. The backend handles data management, receives requests from the frontend and processes them. Simulation data and results are stored persistently in a MongoDB database. EnergyPlus is used for simulation while its integration is handled via the python package eppy. In the following, we describe the backend and frontend components. For more information, we refer to the documentation page[‡].

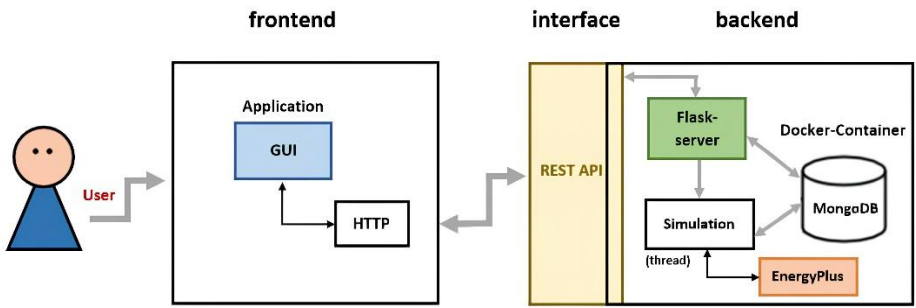

**Fig. 2.** Architecture of EnergyPlus Room Simulator.

### 4.1 Backend

The backend of the system comprises several components. One part handles the MongoDB database, where simulation parameters, such as room model or weather file, are stored and can be accessed later for review. The second part of the backend is the REST API, which manages all interactions with the frontend. On the one hand, it serves as an interface between the two components. On the other hand, it can be used by any application written by the user and enables automation. The backend also allows running multiple simulations with various room sizes, orientations, and infiltration rates successively. The third part of the backend is the simulation control, which comprises pre- and post-processing methods.
Before running a simulation, the system constructs an idf model to define the simulation parameters for EnergyPlus. This model is based on a selected initial idf file, which is then modified according to parameters provided by the user. We consider a rectangular room with a single exterior wall, which is a simple but typical type of room found in numerous buildings. The user defines parameters such as geometry, orientation, and infiltration rate of the room. Windows are adjusted automatically to match the geometric conditions. This means that when room dimensions are changed, the system will keep the original window size from the provided idf file and place as many windows as possible within the available space on the exterior wall. Then, occupancy and ventilation schedules are created in accordance with the user input, and the time range for the desired simulation is defined.

---

[‡] https://ccwi.github.io/EP-Room-Simulator/

Note that the current implementation focuses on room size and presence of occupants, whereas building controls, such as window shading or temperature setpoints, must be defined in advance via the provided initial idf file. After initialization, the backend triggers an EnergyPlus simulation.

### 4.2 Occupancy Scheduling

One of the key functionalities of the backend is the translation from occupancy or window opening data from a sequential tabular (csv) format, common in data science, to its EnergyPlus representation consisting of year, week, and day schedules. The daily schedule object contains the activity data for one day. It is then connected to a weekday in the week schedule. The weeks are further composed in the year schedule where the dates are defined. EnergyPlus Room Simulator defines the activity for each day individually based on the date and assigns it to the correct weekday in the correct week schedule. Thus, it allows using real occupant data or generated occupancy time series instead of repeating design schedules typically used in energy simulations. To generate tabular occupancy data, users may, for instance, use the simulator of Chen et al. [5].

### 4.3 REST API

The REST API offers different endpoints for simulation management. One allows the creation of a new simulation. Simulation metadata entries are accessible via POST and GET requests. Another endpoint starts the simulation. After completion, the results are provided in the EnergyPlus standard output (eso) format and in csv format. A simulation history of previously run simulations can be requested and simulations may be re-opened and re-run with certain modifications. Furthermore, the user may run a simulation series, either by manual programming or by using a provided endpoint that runs several simulations.

### 4.4 Frontend

The frontend offers to run and configure simulations via an easy-to-use GUI. Most of the functionality provided by the REST API can be used via the frontend as well. Initialization, simulation, and result viewing follow a clear process, displayed as a navigation bar (see Figure 1). First, the required files (basic room model and weather file) are uploaded, the time range for the simulation is set, and occupancy is defined manually or uploaded via a csv file. Subsequently, room parameters are adjusted in the process step "Room" before starting the simulation. The room model can be visualized in 3D. This visualization is enabled by the integration of the open-source project Ladybug Spider [8]. The Ladybug Spider IDF viewer is integrated as a local component into the project and is run on its own node.js server.

After completion of a simulation, the results may either be downloaded or customized plots may be created for the following factors:

- Simulated zone air temperature
- Simulated zone $CO_2$ concentration
- Simulated zone relative humidity
- Outdoor temperature
- Outdoor air pressure
- Occupancy state
- Window state

Old simulation results can be re-opened, re-downloaded, and re-visualized in plots.

## 5 Conclusion

In this paper, we introduced our tool *EnergyPlus Room Simulator*. It is intended to support research in areas like building automation or energy performance optimization. Therefore, it enables the simulation of environmental data related to a building room, which would otherwise require empirical measurement. This is especially useful in applications of deep learning, which are typically data-intensive. The source code is publicly available and may be further developed.

## Acknowledgement

The work was majorly supported by several student projects from between 2021 and 2023 at the Munich University of applied sciences.

## References

1. IEA (International Energy Agency), World Energy Outlook 2023. (2023) Retrieved 2024-03-19 from https://iea.org/reports/world-energy-outlook-2023
2. M. Weber, C. Doblander, and P. Mandl, Detecting Building Occupancy with Synthetic Environmental Data, in Proceedings of the BuildSys '20 conference, Virtual Event, Yokohama, Japan, November 16-19 (2020), 324-325. https://doi.org/10.1145/3408308.3431124
3. M. Weber, F. Banihashemi, P. Mandl, H.-A. Jacobsen, and R. Mayer, Overcoming Data Scarcity through Transfer Learning in $CO_2$-Based Building Occupancy Detection, in Proceedings of the BuildSys '23 conference, Istanbul, Turkey, November 15-16 (2023), 1-10. https://doi.org/10.1145/3600100.3623718
4. F., Banihashemi, M. Weber, and W. Lang, Deep learning for predictive window operation modeling in open-plan offices. Energy and Buildings **310**, 114109 (2023) https://doi.org/10.1016/j.enbuild.2024.114109
5. Y. Chen, T. Hong, and X. Luo, An agent-based stochastic Occupancy Simulator. Build. Simul. **11**, 37–49 (2018). https://doi.org/10.1007/s12273-017-0379-7
6. T. Zhang, O. Ardakanian, COBS: COmprehensive Building Simulator, in Proceedings of the BuildSys '20 conference, Virtual Event, Yokohama, Japan, November 16-19 (2020), 314–315. https://doi.org/10.1145/3408308.3431119
7. G. Chaudhary, H. Johra, L. Georges, and B. Austbø, Synconn_build: A python based synthetic dataset generator for testing and validating control-oriented neural networks for building dynamics prediction. MethodsX **11**, 102464 (2023) https://doi.org/10.1016/j.mex.2023.102464
8. Ladybug Tools LLC, Spider. (2018) Retrieved 2024-03-19 from https://www.ladybug.tools/spider